\documentclass{isrpaper}

\usepackage{fontspec}

\usepackage{orcidlink}

\usepackage{etoolbox}
\AtBeginEnvironment{tabular}{\scriptsize}
\usepackage[round,authoryear]{natbib}

\newcommand{\parencite}[1]{\citep{#1}}
\newcommand{\textcite}[1]{\citet{#1}}

\title{Biomechanical 3D Body: Self-Supervised Distillation of Biomechanical Pose from a 3D Body Foundation Model}

\author{
R. James Cotton\,\orcidlink{0000-0001-5714-1400}\,$^{1,2}$, J.D. Peiffer\,\orcidlink{0000-0003-2382-8065}\,$^{1,3}$, Lucinda Williamson$^{4}$, John Leske$^{1,3}$, Georgios Pavlakos$^{5}$}

\affiliations{
$^{1}$Shirley Ryan AbilityLab, Chicago, IL, USA \\ $^{2}$Physical Medicine \& Rehabilitation, Northwestern University, Chicago, IL, USA \\ $^{3}$Biomedical Engineering, Northwestern University, Evanston, IL, USA \\ $^{4}$Biomedical Engineering, University of Illinois Chicago, Chicago, IL, USA \\ $^{5}$Computer Science, University of Texas at Austin, Austin, TX, USA}

\paperdate{August 2026}

\correspondence{rcotton@sralab.org}

\paperabstract{State-of-the-art monocular body recovery methods predict mesh vertices and angles on the corresponding kinematic tree, but their outputs lack biomechanically defined joint angles that downstream applications like clinical and biomechanical analyses require. We extend an existing foundation model, SAM-3D-Body, with an additional biomechanical prediction head that, from a single RGB image, regresses the joint angles and scales of a biomechanical model. Training this model presents a challenge, as there are limited datasets of paired images and biomechanical fits. To overcome this, we supervise biomechanical outputs with in-loop optimized targets from a Levenberg-Marquardt solver performing inverse kinematics fits against markers from the mesh predictions. This allows distilling the biomechanical head from the mesh head, even from unlabeled images. To make this work with GPU-optimized biomechanical models in MuJoCo, the entire model was implemented in JAX using Equinox. We trained this distilled output head on the publicly released SAM-3D-Body dataset. We then validated this model on biomechanical fits to two publicly available marker-based datasets, MoVi and BioCV, as well as movements from a clinical cohort captured with multiview markerless motion capture. The resulting model outperforms existing models for direct regression of biomechanics from images while only slightly underperforming the state-of-the-art monocular biomechanics method that performs more costly inference-time optimization of entire trajectories.}

\begin{document}

\maketitle

\section{Introduction}

Movement directly reflects neurological and musculoskeletal health, yet it is rarely measured in routine clinical care. Multiview markerless motion capture (MMMC) has matured to the point where, on well-observed movements, it agrees closely with marker-based optical systems for gait and upper-limb analysis \parencite{kanko_concurrent_2021, unger_differentiable_2025, cotton_differentiable_2025, firouzabadi_biomechanical_2024}. This progress rides on the rapid evolution of monocular human pose estimation, which now culminates in large foundation models that accurately estimate 2D and 3D keypoints \parencite{sarandi_learning_2023, khirodkar_sapiens_2024, khirodkar_sapiens2_2026}. These keypoint trajectories can then be used to recover biomechanically defined joint angles following the ISB standards for the lower and upper limb \parencite{wu2002isb, wu2005isb}, either by inverse kinematics (IK) on triangulated virtual keypoint trajectories \parencite{uhlrich2023opencap} or by directly optimizing biomechanical fits to minimize reprojection error \parencite{cotton_differentiable_2025, donahue_calibrated_2026}.

In parallel, there has been substantial progress in estimating high-quality 3D mesh representations of humans from monocular images \parencite{kanazawa_end--end_2018, shin_wham_2024, yang_sam_2026}. SAM-3D-Body is a particularly powerful recent state-of-the-art method that predicts the parameters of the Momentum Human Rig (MHR) \parencite{ferguson_mhr_2025}, a model that decouples a posable skeleton from a shape-dependent surface. However, the kinematic trees of MHR and other widely used mesh models such as SMPL do not directly respect the anatomical joint definitions or rotational degrees of freedom required for clinical and biomechanical analyses \parencite{loper_smpl_2015}. This mismatch has motivated recent work on models such as SKEL, which incorporates skeletal bones into a surface mesh representation and begins to bridge computer-vision body models with biomechanical analysis \parencite{keller_osso_2022, keller2023skel}.

Several recent monocular systems have built on these advances to estimate biomechanical quantities from video. Human Skeleton and Mesh Recovery (HSMR) \parencite{xia_reconstructing_2025}, for example, uses the SKEL model but addresses the lack of paired image/SKEL labels by converting SMPL pseudo-ground-truth from existing pose datasets into SKEL parameters. It then refines these labels during training with SKELify, a SMPLify-style iterative optimizer that fits SKEL parameters to 2D keypoints. HSMR is optimized for keypoint position accuracy, which is an important computer-vision metric but does not reliably predict joint-angle accuracy \parencite{wang_mpjpe_2026}. Recently, OpenCap released a monocular version of its software \parencite{gilon_opencap_2026} that computes IK fits to SMPL estimates from WHAM \parencite{shin_wham_2024} with additional physics-constrained refinement. Finally, our recent Portable Biomechanics Laboratory (PBL) \parencite{peiffer_portable_2026}, uses 2D and 3D keypoint locations from an existing foundation model \parencite{sarandi_learning_2023} and optimizes a biomechanical fit to these data by minimizing reprojection error and a 3D keypoint loss while enforcing a single skeleton scaling across multiple trials. Although PBL reports state-of-the-art accuracy and was validated against both marker-based and MMMC in a large, clinically representative population, its optimization-based fitting process remains relatively slow.

This motivated us to investigate whether we could adapt a state-of-the-art vision model for estimating mesh representations from video, SAM-3D-Body, to output biomechanical estimates in a feed-forward manner while achieving accuracy comparable to optimization-based approaches. Two obstacles stand between a mesh foundation model and feed-forward biomechanics. First, there is no large \textit{paired} dataset of images annotated with biomechanical ground truth, so a regressor cannot simply be trained against joint-angle labels. We recently showed that accurate biomechanical estimates can be obtained from SAM-3D-Body MHR estimates using IK fits to individual MHR poses \parencite{cotton_monocular_fingers_2026}, similarly to SKELify. Second, popular biomechanical simulators such as OpenSim \parencite{delp_opensim_2007} and Nimblephysics/AddBiomechanics \parencite{werling_addbiomechanics_2023} are not built to integrate into model training. MuJoCo \parencite{todorov_mujoco_2012}, through its JAX accelerated MJX backend, makes forward kinematics (FK) differentiable and highly performant, but using it would require linking the Torch-based SAM-3D-Body model to a JAX-based biomechanical model.

Our approach addresses both obstacles at once. We port SAM-3D-Body and MHR to JAX/Equinox \parencite{jax_2018, kidger_equinox_2021} so that the frozen foundation model and the differentiable MJX FK live in a single differentiable framework. Onto the \textit{frozen} backbone and decoder we graft a small trainable biomechanical head that reads the existing decoder tokens and regresses the generalized coordinates (qpos), per-group segment scales, and anatomical marker offsets of a MuJoCo biomechanical model. To address the absence of paired labels, we draw on model-in-the-loop optimization (SPIN) \parencite{kolotouros_spin_2019}, which during training optimizes a model's outputs to better align with a more accurate teacher, mirroring the refinement HSMR obtains from 2D keypoints with SKELify. Concretely, a loss function quantifies MHR-to-biomechanics alignment by enforcing pose similarity between site-marker positions (computed via differentiable MuJoCo FK) and MHR-derived markers, requiring no ground-truth biomechanical labels. An inline Levenberg--Marquardt IK solver optimizes this objective in the training loop and provides pseudo-labels for qpos and scale. We show the resulting model, which we term Biomechanical 3D Body or B3D for short, is competitive with SOTA monocular biomechanical methods that utilize extensive optimization.

This paper makes three contributions:

\begin{enumerate}
\item \textbf{Dual-head foundation model.} A small trainable biomechanical head on a \textit{frozen} SAM-3D-Body backbone adds clinically interpretable, MuJoCo-compatible biomechanical outputs with minimal parameter and compute cost.
\item \textbf{Label-free distillation.} The head is trained entirely from the frozen MHR head through an expectation-maximization (EM)-fitted marker mapping and differentiable MuJoCo FK, requiring no paired image--biomechanics dataset. While this is trained on a dataset with ground truth MHR pseudo-labels, these labels are not used in training, only the SAM-3D-Body predictions.
\item \textbf{Model-optimization-in-the-loop biomechanical training.} An inline Levenberg--Marquardt IK solver supplies cold-to-warm pseudo-labels and, by learning to seed it, the head drives the IK past what cold starts reach.
\end{enumerate}

\section{Related Work}

\textbf{Surface-mesh body recovery.} Many monocular 3D body methods recover a surface
mesh, with SMPL \parencite{loper_smpl_2015} parameterizing the body surface by pose and
shape. SMPLify \parencite{bogo_keep_2016} demonstrated fitting SMPL meshes to images through
optimization against 2D keypoints constrained by a plausible human pose prior.
Human mesh recovery (HMR) then showed a model could be trained to directly regress
those parameters from an image \parencite{kanazawa_end--end_2018}.
SPIN \parencite{kolotouros_spin_2019} combined these methods to run an SMPLify-style optimizer \textit{inside} the
training loop to produce refined pseudo-labels that supervise the regressor, improving performance without full 3D ground truth.
Recently, SAM-3D-Body \parencite{yang_sam_2026} trained a state-of-the-art foundation model for mesh recovery on a massive labeled dataset
using a DINOv3 backbone \parencite{simeoni_dinov3_2025} and an MHR output head \parencite{ferguson_mhr_2025}, where MHR inherits the skeleton/shape decoupling of ATLAS \parencite{park_atlas_2025}. All these emit a deformable surface
rather than the biomechanical joint angles and segment lengths this work targets.

\textbf{Markerless biomechanical estimators.}
OSSO \parencite{keller_osso_2022} learned from DXA scans to infer an internal
skeleton from the body surface, and SKEL \parencite{keller2023skel} then re-rigged
the SMPL body with a biomechanically accurate skeleton --- optimized inside
registered SMPL meshes --- yielding a skinned skeleton model whose joints follow
biomechanical conventions. HSMR
\parencite{xia_reconstructing_2025} leverages SKEL by first generating a
pseudo-labeled training set with SKELify (a SMPLify-style optimizer that fits
SKEL parameters to trustworthy 2D keypoints). Following the SPIN-style approach
described above, it periodically refines these labels during training to learn
a direct regressor from images to SKEL parameters, yielding both a full body
mesh and biomechanically constrained joint angles. Our
method adopts the same in-the-loop refinement, but supervises against a 3D
biomechanical teacher through differentiable simulation rather than 2D keypoints. OpenCap-Monocular
\parencite{gilon_opencap_2026} first recovers a per-frame SMPL body mesh from video with WHAM
\parencite{shin_wham_2024} , then extracts virtual markers from that mesh to scale and fit an OpenSim
\parencite{delp_opensim_2007} musculoskeletal model to the recovered keypoints,
which amounts to an IK step on the WHAM output. Similarly,
BioPose \parencite{koleini_biopose_2025} regresses anatomically constrained joint
angles with a neural IK network over mesh-vertex virtual markers,
but its code and weights were not released, so we do not compare against it. The
Portable Biomechanics Laboratory (PBL) \parencite{peiffer_portable_2026} shows
state-of-the-art performance for recovering biomechanical kinematics from
monocular video by leveraging a vision foundation model for 2D and 3D keypoints
\parencite{sarandi_learning_2023} and then optimizes kinematic trajectories
end-to-end to minimize the 3D and reprojection errors; it is the strongest
baseline we compare against here, and its trajectory optimizer also produces the
reference fits for two of our evaluation cohorts. In further prior work we
showed that accurate biomechanical joint angles, including individual
fingers, can be recovered from MHR meshes by minimizing a pose-similarity
distance with a Levenberg--Marquardt IK optimizer
\parencite{cotton_monocular_fingers_2026}; that solver is the one we run inside the
training loop here.

\section{Methods}

\begin{figure}[!htbp]
\centering
\includegraphics[width=0.85\linewidth]{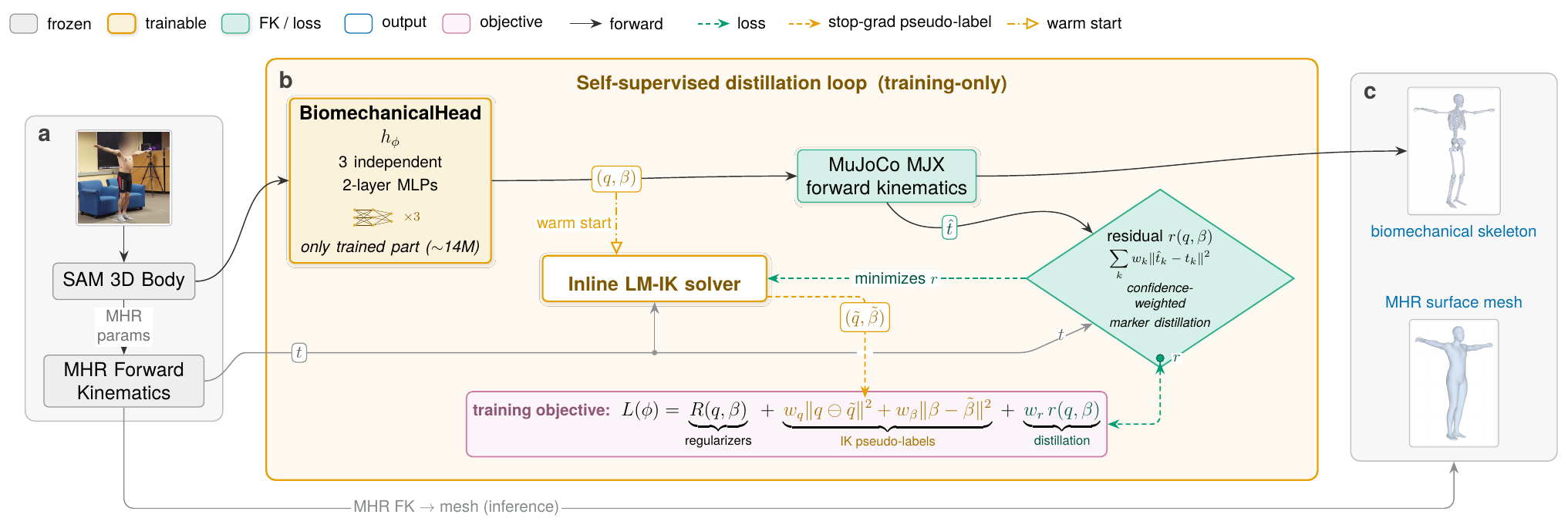}
\caption[]{Method overview. A person crop drives the frozen SAM-3D-Body backbone to a pose
token, read by the frozen MHR head and trainable biomechanical head. MJX FK turns the predicted parameters into site positions, matched to
MHR-derived targets via a confidence-weighted loss. An inline SPIN-style IK
solver supplies warm-started pseudo-labels. At inference a single feed-forward
pass emits both the MHR mesh and the biomechanical skeleton.}
\label{fig-architecture}
\end{figure}

\textbf{Overview.}
We graft a small trainable \textit{biomechanical head} onto a frozen SAM-3D-Body backbone \parencite{yang_sam_2026} and train it
to regress the joint angles and body scale parameters for a MuJoCo-MJX biomechanical model \parencite{todorov_mujoco_2012} by
distilling from the frozen MHR head
\parencite{ferguson_mhr_2025}. To overcome the lack of paired image-to-biomechanics labels, the
biomechanical head's outputs are pushed through differentiable MuJoCo forward FK and
matched to MHR-derived targets. A SPIN-style \parencite{kolotouros_spin_2019} optimization in the training loop is performed using a
Levenberg--Marquardt \parencite{levenberg_method_1944, marquardt_algorithm_1963} IK solver to supply pseudo-labels. The entire model is implemented in JAX/Equinox \parencite{jax_2018, kidger_equinox_2021} so gradients flow from image features through to the biomechanical model running in MJX. Figure~\ref{fig-architecture} summarizes the architecture.

\subsection{Biomechanical model}

For the biomechanical model, we use a full-body MuJoCo humanoid with a flexible lumbar spine,
ball-and-socket hips and shoulders, and fully articulated hands (99 generalized
coordinates). It carries 102 anatomical marker sites (87 body, 10 fingertip, 5 face),
rigged to the scaled model via per-marker offsets that capture individualized
surface geometry.
We disabled knee internal/external rotation, varus/valgus, and metatarsophalangeal flexion. We exclude finger pose from validation.

FK is implemented by MJX, a differentiable MuJoCo simulator running on the GPU.
Let $q$ be the generalized coordinates and, following the SMPL-style $\mathcal{M}(\theta,\beta)$ convention \parencite{loper_smpl_2015}, $\beta \in \mathbb{R}^{14 + 102\times 3}$ the model's shape parameters --- the 14 per-segment scale parameters concatenated with the 102 per-marker offsets. We write the model scaling
together with its MJX FK as a single differentiable operator
$\mathcal{B}$ that returns the world-frame joint centers $J$ and the site-marker
positions $x$:

\begin{equation}
\label{eq:biomech}
(J,\, x) = \mathcal{B}(q,\, \beta).
\end{equation}

Although we write $\mathcal{B}(q,\beta)$ by analogy with $\mathcal{M}(\theta,\beta)$, the
parameters mean different things. In a SMPL model the pose $\theta$ is a set of
per-joint rotations of a tree fit to registered scans and the shape $\beta$ is a
coefficient vector in a learned basis, so joint locations and segment lengths are
implied by the surface.
Here $q$ is the generalized coordinates of a biomechanical model, and thus are inherently clinically defined and interpretable.
Correspondingly $\beta$ is anatomical rather than statistical: per-segment scale factors
and per-marker offsets, so segment lengths are explicit and individually scaled.

\subsection{MHR Model}

MHR is a parametric body model \parencite{ferguson_mhr_2025}
whose posable kinematic skeleton is decoupled from a shape-dependent surface, unlike SMPL. While the scale parameters are more directly interpretable than SMPL $\beta$ parameters, the kinematic tree and joint angles are not biomechanically defined.  Writing its pose and shape
parameters as $\theta_\mathrm{M}$ and $\beta_\mathrm{M}$, MHR provides posed mesh
vertices $V$, regressed keypoints $K$, and kinematic
joint centers $G$:

\begin{equation}
\label{eq:mhr}
(V,\, K,\, G) = \mathcal{M}(\theta_\mathrm{M},\, \beta_\mathrm{M}).
\end{equation}

\subsection{MHR-to-biomechanical marker mapping}

The MHR and MuJoCo biomechanical models do not share marker sites, joint definitions, or generalized coordinates, so we first construct a correspondence between their observable 3D locations. We use the BML-MoVi marker set as the target marker layout because it provides a dense set of 87 anatomical body markers that we have shown to support accurate joint-angle estimation compared with sparser marker sets \parencite{ghorbani_movi_2021, cotton_optimizing_2023, cotton_differentiable_2025, peiffer_portable_2026}.

For each biomechanical marker, we assign a corresponding MHR output, similar to our recent work \parencite{cotton_monocular_fingers_2026}. Some assignments are direct: for example, kinematic joint centers such as the hip or knee are closely matched to the corresponding MHR joint centers. Surface markers are less obvious, and for these markers we estimate correspondences between MoVi marker locations and SAM-3D-Body mesh vertices by comparing their image-plane projections across many video sequences, weighting candidate matches by reprojection agreement. This produces a fixed mapping in which each biomechanical marker corresponds to one MHR output: either a kinematic joint center, a mesh-regressed keypoint, or a mesh vertex.

We then refine the mapping with an offline EM procedure on the SAM-3D-Body dataset \parencite{yang_sam_2026}. In the E-step, the current MHR targets are fit with IK over the biomechanical model $\mathcal{B}$, solving for joint angles $q$ while sharing body scale and marker offsets across frames from the same subject. In the M-step, the pooled residuals are used to update the population-level marker template and a confidence weight $w_k$ for each marker. Markers with small, consistent residuals are assigned higher confidence weights, whereas markers with noisier correspondences are down-weighted.

The final correspondence is represented by two gather operators that select matched points from each model in the same marker order. The biomechanical prediction, $\hat t(q,\beta) = \Pi_\mathrm{B}(J, x)$, selects the corresponding marker locations from the biomechanical joint centers and site positions returned by Eq.~\ref{eq:biomech}. The MHR target, $t = \Pi_\mathrm{M}(V, K, G)$, selects the matched vertices, keypoints, or joint centers from the MHR outputs in Eq.~\ref{eq:mhr}. The IK solver then minimizes the confidence-weighted distillation residual

\begin{equation}
\label{eq:distill}
r(q,\beta) = \sum_{k=1}^{N} w_k\,\lVert \hat t_k(q,\beta) - t_k\rVert^2 ,
\end{equation}

where $N$ is the number of matched markers and $w_k$ are the marker-specific confidence weights learned by the EM procedure.

\subsection{Dual-head grafting}

The biomechanical head reads the frozen decoder's 1024-d pose token concatenated with the frozen MHR head's output (body pose and camera) projected to token space (2048-d input), through three parallel two-layer MLPs (hidden width 2048, ReLU), one per output: generalized coordinates (6D continuous rotations \parencite{zhou_6d_2019} for the root and ball joints, mapped to unit quaternions by Gram--Schmidt), per-segment body scales, and per-marker offsets. This adds {\textasciitilde}13.6 M trainable parameters while keeping perception and the MHR head frozen (full dimensions in Appendix A.12).

\subsection{Training objective and inline Levenberg--Marquardt IK}

During training we obtain pseudo-labels from a Levenberg--Marquardt IK solve run every step, mirroring SPIN's inner optimizer \parencite{kolotouros_spin_2019}. The Levenberg--Marquardt IK minimizes $r$ Eq.~\ref{eq:distill} over $(q,\beta)$ in staged fashion (root, then full pose, then scale and offsets), following our prior work \parencite{cotton_monocular_fingers_2026}. It is seeded from the neutral pose and, as the head improves, warm-started from its biomechanical predictions; the solution $(\tilde q,\tilde\beta)$ is stop-gradiented before use as a target.

The biomechanical head is trained to minimize a single self-supervised objective built on the
distillation residual $r$ of Eq.~\ref{eq:distill}. Let $(q,\beta)=h_\phi(z)$ be the head's
prediction from the pose token $z$, the training objective is

\begin{equation}
\label{eq:loss}
\mathcal{L}(\phi) = \underbrace{\textcolor{teal}{w_\text{r}\, r(q,\beta)}}_{\textcolor{teal}{\text{distillation}}}
  \;+\; \underbrace{\textcolor{orange}{w_q\,\lVert q \ominus \tilde q\rVert^2 + w_\beta\,\lVert \beta - \tilde\beta\rVert^2}}_{\textcolor{orange}{\text{IK pseudo-labels}}}
  \;+\; \underbrace{\mathcal{R}(q,\beta)}_{\text{regularizers}}.
\end{equation}

The distillation term backpropagates $r$ through the differentiable FK of
$\mathcal{B}$ into the head, requiring no biomechanical ground truth; the
pseudo-label terms pull the same prediction toward the inline-IK optimum. Here
$\ominus$ is the Euclidean difference on translational/hinge coordinates and a
quaternion distance on the orientation coordinates (root and ball joints).
$\mathcal{R}(q,\beta)$ collects the anatomical regularizers (joint-limit and
ball-cone penalties, scale and quaternion normalization priors; see Appendix A.8).

\subsection{Datasets, augmentation, and metrics}

We train on the public SAM-3D-Body dataset \parencite{yang_sam_2026, sam3dbody_dataset_hf}
--- \textbf{1,486,720 distinct images (3,722,776 person-crop training samples)} drawn from
five of its constituent image sources --- with mild augmentation (color and
bounding-box jitter; no flips or rotations); see Appendix A.7 for the
per-source breakdown. We evaluate on \textbf{entirely separate biomechanical datasets} to test generalization. First, BML-MoVi (85
participants) provides marker-based tracking with synchronized monocular videos; we represent the reference kinematic trajectory as a
learned implicit function following \parencite{peiffer_portable_2026} and score monocular inference against it. Second, because public
datasets under-represent movement impairments --- a key motivation here --- we evaluate on synchronized MMMC data with
monocular smartphone videos, 1693 trials from 109 participants across four clinical populations (controls, lower-limb prosthesis users, participants with
neurological injury, and pediatric inpatients). Third, we add the BioCV dataset \parencite{evans_biocv_2024} --- 308 trials from 14 participants performing walking, running, hopping, and both self-selected and maximal effort counter-movement jumps --- a
second marker-based reference, with monocular inference scored against a MuJoCo IK reconstruction (the optical-mocap gold standard). 308 trials remain after quality gating of the 362 recordings (static poses, non-converged
reference fits, and un-trackable clips removed; details in Appendix A.10.1). The feed-forward regressors (ours, HSMR)
run per-frame; the optimization methods (PBL, OpenCap) fit the full video. OpenCap has additional filtering based on keypoint detection thresholds that
causes it to drop some trials entirely (see Appendix A.10).

For joint angles on each trial we measure the median joint-angle error (MJAE), the median over
frames of the absolute signed error in degrees. We report the cohort statistic as the
median of per-trial MJAE and with normalized inter-quartile range (nIQR).
For joint centers we report the Procrustes-aligned mean per-joint position error
(PA-MPJPE), the mean Euclidean residual after a per-frame similarity alignment
(scale, rotation, translation). PA-MPJPE is computed over the 19 common lower- and
upper-limb joint centers shared by all methods (torso, head, and the dense finger
joints excluded; full list in Appendix A.10).

\subsection{Inference and IK refinement}

At inference our method is a stateless per-frame forward pass on the evaluation crops to $(q,\beta)$, with no temporal window (unlike PBL and OpenCap). Because distillation does not fully reach the warm-IK quality seen in training, we optionally apply the same warm Levenberg--Marquardt-IK refinement at inference and report results with and without it.

\subsection{Comparisons}

We compare against the Portable Biomechanics Laboratory (PBL)
\parencite{peiffer_portable_2026}, HSMR \parencite{xia_reconstructing_2025}, and
OpenCap Monocular \parencite{gilon_opencap_2026} --- the existing monocular-video
biomechanics methods with released code and weights. Each method and reference
uses a different biomechanical skeleton definition, which we reconcile using ISB
conventions \parencite{wu2002isb, wu2005isb}. A single fixed per-(method, joint) offset,
pooled across the whole cohort, then removes the residual constant cross-skeleton
convention difference without absorbing genuine per-subject pose error. An angle is
compared only where the parameterizations line up --- the joint is single-DOF or we read
the first axis of a matching Cardan sequence, and the sign has been resolved from each
model's own live joint axes --- which covers hip, knee, ankle and elbow flexion, while
joints that differ in kind, such as the lumbar spine, are not compared.

\section{Results}

We compare our monocular method against three baselines (HSMR, OpenCap-Monocular,
and the optimization-based PBL) on three datasets. To render metrics equivalent across the slightly different skeleton
formats and avoid penalizing the non-MuJoCo models for rigging differences, we report
bias-corrected metrics alongside raw joint-angle metrics, subtracting a single per-joint
median offset (pooled over all trials against the reference) from each method before scoring.

\subsection{Qualitative comparison}

\begin{figure}[!htbp]
\centering
\includegraphics[width=0.72\linewidth]{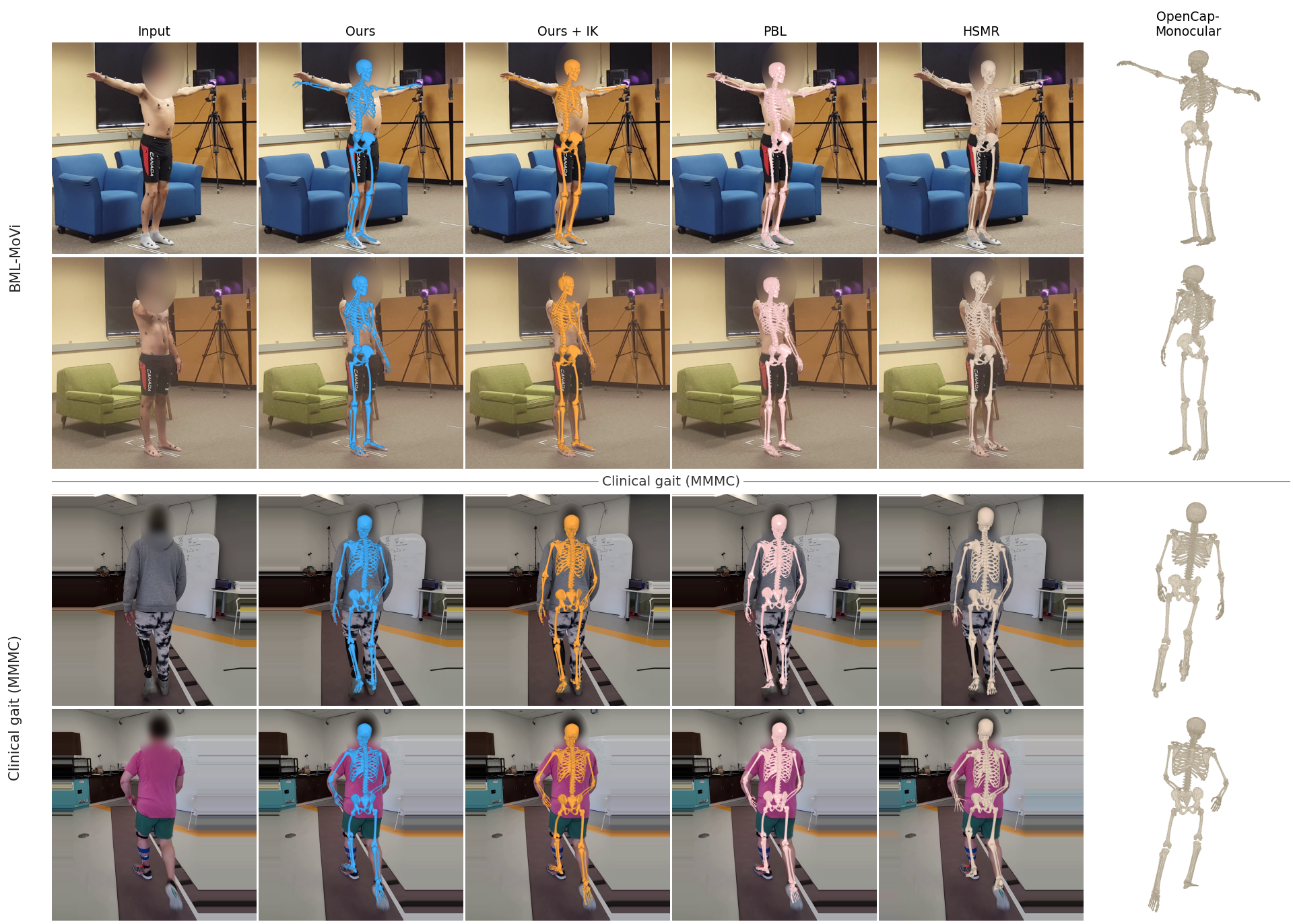}
\caption[]{Qualitative overlays on BML-MoVi (top two rows) and MMMC clinical data (bottom
two rows). OpenCap-Monocular does not provide overlay code.}
\label{fig-qualitative}
\end{figure}

\begin{figure}[!htbp]
\centering
\includegraphics[width=0.72\linewidth]{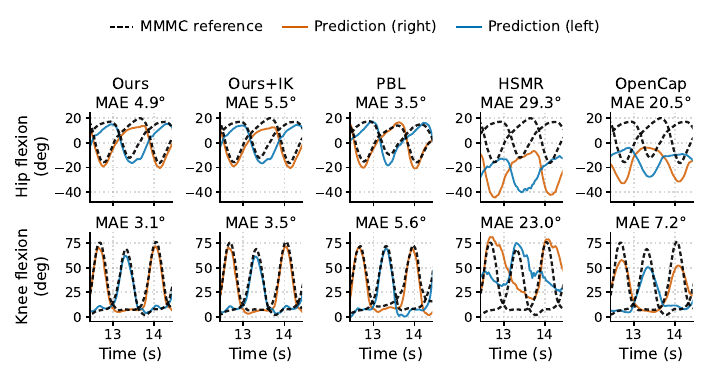}
\caption[]{Per-method gait waveforms on one clinical walking trial from a participant with no
movement impairment. Top: hip flexion; bottom: knee flexion. Predictions are shown
\textbf{raw (without bias correction)} so each method's native joint-angle offset is
visible; the per-panel MAE is therefore uncorrected and computed over this
two-second window alone, and is not comparable to the cohort metrics in
Table~\ref{tab-mmc-metrics}.}
\label{fig-gait-waveforms}
\end{figure}

Figure~\ref{fig-qualitative} shows qualitative results from each method on frames from the BML-MoVi and MMMC datasets. While all methods perform fairly well most of the time, our method shows slightly better alignment than HSMR in places (e.g., feet); OpenCap-Monocular provides no overlay code, so it cannot be shown. Kinematic waveforms are shown in Figure~\ref{fig-gait-waveforms} on an example walking trial, with our method and PBL track the reference more tightly than HSMR and OpenCap.

\subsection{Accuracy against marker-based references (BML-MoVi and BioCV)}

Both BML-MoVi and BioCV provide synchronized marker-based motion capture and monocular video, though each uses a different reference-fitting approach: BML-MoVi's reference kinematics are reconstructed from the markers as a learned implicit function \parencite{peiffer_portable_2026}, while BioCV's reference is a marker-based MuJoCo IK reconstruction. Monocular inference is scored against each. The clinical cohort (next section) is referenced against MMMC.

Our method tracks the dominant lower-limb angles
tightly and recovers joint centers to within several centimetres on both references
(Table~\ref{tab-marker-metrics}). It is competitive with the optimization method
PBL --- which reports current SOTA performance for monocular biomechanics --- and
outperforms both HSMR and OpenCap-Monocular across all metrics.
Figure~\ref{fig-error-boxplots}a breaks the BML-MoVi results out by individual
joint, where these trends hold. On BML-MoVi, enabling the optional inline IK
at inference had minimal impact.

\begin{table}
\centering
\caption[]{Cross-method comparison against the two marker-based references (BML-MoVi optical mocap; BioCV marker-based MuJoCo IK). Rows are median (nIQR) bias-corrected joint-angle error (degrees) pooled per limb, and PA-MPJPE (mm) over the common joint centers. Best per row in \textbf{bold}; lower is better. \textbf{†} marks the optimization-with-kinematic-fitting methods (PBL, OpenCap), vs the feed-forward regressors.}

\begin{tabular}{lccccc}
\toprule
Metric & Ours & Ours+IK & PBL$^\dagger$ & HSMR & OpenCap$^\dagger$ \\
\midrule
\multicolumn{6}{l}{\textit{BML-MoVi (optical mocap, 85 subjects)}} \\
Lower-limb & 3.25 (1.38) & 3.15 (1.43) & \textbf{2.73 (1.29)} & 4.76 (3.81) & 3.84 (2.86) \\
Upper-limb & 6.44 (5.57) & \textbf{6.31 (5.52)} & 7.25 (6.41) & 12.05 (9.54) & 10.31 (5.97) \\
PA-MPJPE & 37.1 & 36.0 & \textbf{33.9} & 49.8 & 47.2 \\
\midrule
\multicolumn{6}{l}{\textit{BioCV (marker-based MuJoCo IK, 14 subjects, 308 trials)}} \\
Lower-limb & 4.04 (2.15) & \textbf{3.96 (2.29)} & 4.10 (2.43) & 6.11 (4.12) & 5.39 (3.88) \\
Upper-limb & \textbf{6.25 (6.38)} & 6.36 (6.05) & 6.69 (6.91) & 12.05 (12.57) & 10.35 (9.98) \\
PA-MPJPE & 62.2 & 62.7 & \textbf{52.5} & 67.4 & 64.9 \\
\bottomrule
\end{tabular}
\label{tab-marker-metrics}
\end{table}

\subsection{Accuracy on the clinical MMMC cohort}

\begin{figure}[!htbp]
\centering
\includegraphics[width=0.74\linewidth]{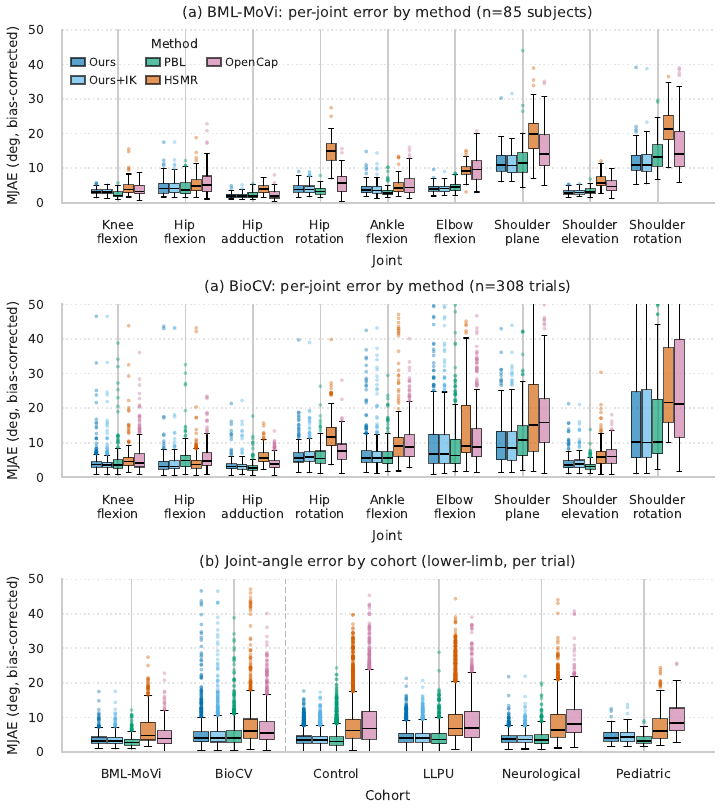}
\caption[]{Bias-corrected joint-angle error (MJAE, deg) across methods. \textbf{(a)} Per-subject
error by joint on the external BML-MoVi cohort. \textbf{(b)} Per-trial lower-limb error by cohort ---
BML-MoVi (leftmost, left of the dashed guide), the four clinical gait
populations, and the BioCV cohort --- vs their respective
references. Our method stays flat from MoVi through
the impairment cohorts while HSMR inflates; PBL is tightest on MoVi and the clinical
populations. OpenCap appears where
its front-end produced a fit.}
\label{fig-error-boxplots}
\end{figure}

We evaluated on a MMMC reference spanning
four clinical populations (controls, lower-limb prosthesis users, participants with
neurological injury, and pediatric inpatients). Table~\ref{tab-mmc-metrics}
reports the same metrics as on the marker-based references and shows the same trend: our method is highly competitive against the other baselines except PBL, which outperforms it. Again, IK during inference had minimal impact.
Figure~\ref{fig-error-boxplots}
breaks down the average joint-angle error by cohort (including the two marker-based
references for comparison), where our advantage over HSMR and OpenCap-Monocular becomes more
pronounced in several groups. OpenCap produced no scored fit for 400 of 1693 inputs --- its front-end input-quality gate rejects the posterior and self-occluded views common here (details in Appendix A.10) --- excluding them from its analysis and possibly biasing its reported accuracy favorably.

\begin{table}
\centering
\caption[]{Cross-method comparison on the clinical gait cohort using the multi-camera markerless reference, pooled across four clinical populations. Columns and metrics as in Table~\ref{tab-marker-metrics}. Best per row in \textbf{bold}; lower is better. \textbf{†} marks the optimization-with-kinematic-fitting methods (PBL, OpenCap), vs the feed-forward regressors; PBL's optimization runs orders of magnitude slower (see Inference speed below). Per-population detail in Figure~\ref{fig-error-boxplots}b.}

\begin{tabular}{p{\dimexpr 0.167\linewidth-2\tabcolsep}p{\dimexpr 0.167\linewidth-2\tabcolsep}p{\dimexpr 0.167\linewidth-2\tabcolsep}p{\dimexpr 0.167\linewidth-2\tabcolsep}p{\dimexpr 0.167\linewidth-2\tabcolsep}p{\dimexpr 0.167\linewidth-2\tabcolsep}}
\toprule
Metric & Ours & Ours+IK & PBL† & HSMR & OpenCap† \\
\hline
Lower-limb & 3.66 (1.61) & 3.59 (1.66) & \textbf{3.24 (1.96)} & 6.36 (4.35) & 7.16 (5.77) \\
Upper-limb & 7.40 (6.99) & 7.44 (6.99) & \textbf{6.18 (5.49)} & 10.45 (10.25) & 11.43 (17.50) \\
PA-MPJPE & 41.9 & 42.0 & \textbf{27.6} & 75.8 & 82.4 \\
Rejected & 0 & 0 & 0 & 0 & 400 \\
\bottomrule
\end{tabular}
\label{tab-mmc-metrics}
\end{table}

\subsection{Training dynamics and fine tuning the MHR head}

\begin{figure}[!htbp]
\centering
\includegraphics[width=0.76\linewidth]{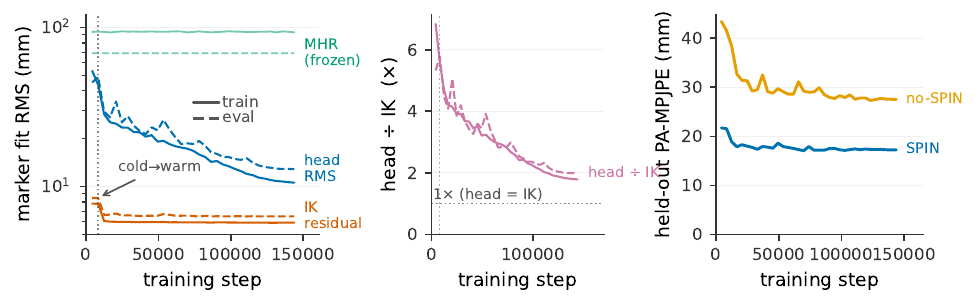}
\caption[]{Training dynamics (train solid, eval dashed) and the SPIN ablation. \textbf{(a)} The IK
marker-fit residual drops at the cold-to-warm transition. \textbf{(b)} The head-to-IK error
ratio falls toward 1$\times$ as the feed-forward head approaches the in-the-loop optimizer.
\textbf{(c)} Ablation: held-out biomechanical joint-center PA-MPJPE, full model (SPIN) vs the
offset-regularized SPIN-disabled baseline.}
\label{fig-training-dynamics}
\end{figure}

Two features of training are visible in Figure~\ref{fig-training-dynamics}. The inline
IK marker-fit residual drops sharply at the cold-to-warm transition, when the solver
switches from a neutral seed to a warm start from the head's own prediction (panel a).
As training proceeds the head's own marker
error falls toward this IK floor but does not reach it (panel b), which motivated the
optional IK mode at inference. Separately, fine-tuning the MHR head improved these
in-distribution metrics while the recovered mesh degraded on off-axis views, so we
kept the MHR head frozen throughout (see Appendix). The full
model trained in 134 GPU-hours (5.6 days, 35 epochs) on a single GPU.

\subsection{Inference speed}

Our method is feed-forward: a single pass through the frozen backbone and grafted
head, with an optional lightweight inline Levenberg--Marquardt-IK refinement. Measured warm over the
BML-MoVi and clinical videos, it runs at roughly 15--20 frames/s in
batched mode, dominated by the SAM-3D-Body backbone forward; the biomechanical head
adds negligible cost. PBL, in contrast, is a whole-clip optimization and runs at
roughly 0.03--0.14 frames/s, two to three orders of magnitude slower.

\subsection{Ablation: model-optimization-in-the-loop (SPIN)}

To isolate the contribution of the inline IK, we retrain the model with the SPIN
pseudo-labels disabled ($w_q = w_\beta = 0$ and the inner solve skipped),
so the head is trained on the distillation residual $r$ Eq.~\ref{eq:distill} alone, under
the same schedule and step budget. The pseudo-label helps regularize the
per-marker offsets, so this baseline required additional regularization to prevent divergence. SPIN improves both the marker fit
($21.1 \rightarrow 12.9$ mm) and the biomechanical joint-center PA-MPJPE ($27.5 \rightarrow 17.3$ mm, Figure~\ref{fig-training-dynamics}c).

A complementary ablation removes the biomechanical head entirely rather than the pseudo-labels during training, and instead runs the
same solver cold on the frozen MHR output at inference that is used during training. Median joint errors are
similar, but the cold solve's error tail is much heavier.
Furthermore, the feed-forward head reaches this accuracy without the cost of performing inverse
kinematics at inference --- 41 ms/frame against 180 ms/frame for the cold solve,
benchmarked at a full batch (Appendix A.5).

\section{Discussion}

We show that it is possible to graft a biomechanical head onto the SAM-3D-Body model and train a model to produce high quality biomechanical estimates from monocular video through self-distillation. We tested our new model with extensive validation on external, real-world datasets that provided either marker-based or MMMC reference values. We found it outperforms an existing human skeleton regression model and is competitive with a state-of-the-art monocular biomechanics method that requires more expensive inference-time optimization.

Importantly for our application domain, we compared methods by measuring the joint angle errors as our primary metric in addition to the MPJPE more common in the human pose estimation literature. This revealed substantial limitations from HSMR, which tends to keep many joints overly flexed, and OpenCap. All methods showed worse accuracy in the arms than the legs, with the shoulder plane and rotation being worst estimated. Accuracy is not uniform across the upper limb, however: elbow flexion and shoulder elevation are recovered to within a few degrees, while the shoulder plane and axial rotation are several times worse. Whether that is sufficient depends on the clinical question --- it may support gross upper-limb screening while falling short of what analyses of humeral plane or axial rotation require. Determining exactly why and when the shoulder and arm are less accurate is left to future work. Viewpoint is one candidate, since our earlier monocular work found frontal-plane angles were best recovered from frontal views and sagittal-plane angles from sagittal ones \parencite{peiffer_portable_2026}; there, however, the effect on lower-limb angles was small, so viewpoint alone seems unlikely to account for errors of this size. We also tested generalization to clinical populations with movement impairments, often recorded in the frontal plane: several comparison methods showed substantially higher errors in these cases, while our method remained robust and competitive with the optimization-based PBL.

Central to our method is the in-loop Levenberg--Marquardt IK: this SPIN-style optimization \parencite{kolotouros_spin_2019} let us distill the biomechanical head using none of the SAM-3D-Body labels, suggesting it could refine the model on additional external video datasets where MHR regression remains robust, though this remains to be tested. An additional benefit of directly regressing the biomechanics, beyond faster inference, is that we saw fewer outliers than when cold-starting inverse kinematics on each frame, which avoids some of the difficulties of optimizing inverse kinematics for high-degree-of-freedom models.

The accuracy of our IK fits are bounded by the loss that aligns the biomechanical model to MHR Eq.~\ref{eq:distill}, which we optimized via a large EM over MHR labels from the SAM-3D-Body dataset; it remains an opportunity for improvement, as some bones slightly extend beyond the MHR mesh. Enabling more degrees of freedom, refining the scaling, or adopting a more complete shoulder with scapulothoracic and glenohumeral motion \parencite{holzbaur_model_2005, firouzabadi_biomechanical_2024} could help, ideally guided by high-quality bone-movement annotations such as biplanar fluoroscopy.

We also compared across different biomechanical models and simulators --- SKEL \parencite{keller2023skel} in HSMR \parencite{xia_reconstructing_2025} (which has no biomechanical simulator) and OpenSim \parencite{delp_opensim_2007} in OpenCap \parencite{uhlrich2023opencap} --- and our model has more spine degrees of freedom than the reference model \parencite{peiffer_portable_2026} used to fit MoVi and MMMC.
We converted all models' conventions to ISB standards \parencite{wu2002isb, wu2005isb}, but since our reference values are all MuJoCo fits, HSMR and OpenCap may be disadvantaged; the per-joint bias correction mitigates this, and the joint-angle errors track the PA-MPJPE errors, further supporting our conclusions.
For the same reason PBL's PA-MPJPE is a lower bound on the clinical cohort, where it is
scored within the reference's own skeleton rather than across two.

While our experiment fine-tuning the MHR head alongside the biomechanical head did not show a benefit (Appendix A.4), optimizing this more carefully and training more of the network is a potential future direction. Furthermore, our earlier IK optimization also enabled accurate monocular finger tracking against MMMC \parencite{cotton_monocular_fingers_2026}, so extending our head to the MHR hand regression is a natural next step.

While faster than our optimization-based PBL \parencite{peiffer_portable_2026} the present method did not exceed its accuracy despite a SOTA vision backbone. PBL remains ahead on Procrustes-aligned joint-center error against all three references, and on the lower limb of both BML-MoVi and the clinical cohort; we are ahead on the upper limb of BML-MoVi and of BioCV, with the two level on the BioCV lower limb, which is enough to carry the BML-MoVi joint-angle average. Against the marker-based references the margins in either direction are small enough that the two are best read as comparable; PBL's advantage is clearest on the clinical cohort, and there most of all in the upper limb.

This method also remains a single-image method, ignoring the inter-frame motion, texture, and appearance cues that can improve biomechanical estimates from video; point-tracking, dynamic scene understanding, and powerful video backbones are all promising directions. Because each frame is estimated independently, nothing in the model enforces temporal continuity. PBL gains accuracy in part through bundle adjustment over reprojection errors with the skeleton scale constrained across frames, whereas our method here is computed independently on each frame. However, our regressed $(q,\beta)$ and the SAM-3D-Body mesh estimates could also seed a similar bundle-adjustment refinement.

\section{Conclusion}

Biomechanical 3D Body (B3D) demonstrates that self-distillation using model in the loop optimization can train a biomechanical output head on a SAM 3D Body backbone that provides performance competitive with state-of-the-art optimization-based methods while massively accelerating inference speed.

\section*{Acknowledgments}
\small
Research reported in this publication was supported by the Eunice Kennedy Shriver National Institute Of Child Health \& Human Development of the National Institutes of Health under Award Number R01HD114776. The content is solely the responsibility of the authors and does not necessarily represent the official views of the National Institutes of Health.
\normalsize

{\small
{\footnotesize

}

\setcounter{secnumdepth}{-2}
\setcounter{table}{0}
\setcounter{figure}{0}
\renewcommand{\thetable}{S\arabic{table}}
\renewcommand{\thefigure}{S\arabic{figure}}
\section{Detailed results and qualitative comparisons}

\subsection{A.1 Full per-joint metric tables}

Per-joint breakdowns of the main-paper summary tables are given in Table~\ref{tab-movi-metrics-detail} (held-out BML-MoVi) and Table~\ref{tab-mmc-metrics-detail} (clinical MMMC). Lower is better throughout. Joint-angle cells report MJAE in degrees, with nIQR in parentheses; both position columns, PA-MPJPE and Rigid MPJPE, are in mm.

The two position metrics measure different joint sets. PA-MPJPE is pooled over the 19 joint centers common to all methods (as in the main text), whereas Rigid MPJPE spans each method's full exposed joint set. They are therefore not strictly comparable across methods, and for a method whose extra joints track well its PA-MPJPE can slightly exceed its Rigid MPJPE.

For joint angles, \textbf{Avg MJAE} is bias-corrected and pooled over the DOFs shared by all methods --- the hip, knee, ankle, elbow and shoulder DOFs --- which is the same set on every dataset. The uncorrected row gives each method's raw per-group mean. \textbf{†} marks the optimization-with-kinematic-fitting methods (PBL, OpenCap).

Cohort sizes are 85 subjects (BML-MoVi) and 1693 trials (clinical MMMC); OpenCap is scored only on the 1293 trials its WHAM front-end accepted (see \texttt{Inputs rejected}).

\begin{table}
\centering
\caption[]{Per-joint cross-method comparison on held-out BML-MoVi. Joint-angle cells are MJAE (deg) with nIQR in parentheses; PA-MPJPE / Rigid MPJPE in mm. Best per row in bold; lower is better.}

\begin{tabular}{p{\dimexpr 0.167\linewidth-2\tabcolsep}p{\dimexpr 0.167\linewidth-2\tabcolsep}p{\dimexpr 0.167\linewidth-2\tabcolsep}p{\dimexpr 0.167\linewidth-2\tabcolsep}p{\dimexpr 0.167\linewidth-2\tabcolsep}p{\dimexpr 0.167\linewidth-2\tabcolsep}}
\toprule
Metric & Ours & Ours+IK & PBL† & HSMR & OpenCap† \\
\hline
\textit{Lower-limb joints} &  &  &  &  &  \\
Knee flexion & 3.15 (0.77) & 2.99 (0.87) & \textbf{2.09 (1.10)} & 3.77 (1.73) & 3.22 (1.90) \\
Hip flexion & 4.08 (2.14) & 4.08 (2.06) & \textbf{3.69 (2.42)} & 4.86 (2.37) & 5.05 (3.39) \\
Hip adduction & 1.89 (0.73) & \textbf{1.88 (0.75)} & 1.99 (1.12) & 3.89 (1.53) & 1.98 (1.30) \\
Hip rotation & 3.75 (1.45) & 3.75 (1.45) & \textbf{3.18 (1.30)} & 14.95 (3.57) & 5.64 (3.34) \\
Ankle flexion & 3.70 (1.35) & 3.48 (1.54) & \textbf{2.82 (0.90)} & 4.18 (1.91) & 4.41 (3.04) \\
\textit{Upper-limb joints} &  &  &  &  &  \\
Elbow flexion & \textbf{3.97 (1.12)} & 4.03 (1.15) & 4.42 (1.36) & 9.23 (1.56) & 9.60 (3.94) \\
Shoulder plane & 10.80 (3.45) & \textbf{10.73 (3.76)} & 11.50 (4.16) & 19.86 (5.39) & 13.97 (6.67) \\
Shoulder elevation & \textbf{2.77 (0.84)} & 2.93 (0.88) & 3.14 (1.01) & 5.67 (1.99) & 4.65 (2.17) \\
Shoulder rotation & 10.88 (3.19) & \textbf{10.79 (3.61)} & 13.22 (4.66) & 21.29 (5.21) & 14.12 (7.68) \\
\textit{Aggregate (pooled angle / MPJPE)} &  &  &  &  &  \\
Avg MJAE (°) & 5.00 & \textbf{4.96} & 5.12 & 9.75 & 6.96 \\
Avg MJAE, uncorrected (°) & 7.65 & 7.58 & 7.68 & 14.21 & 8.29 \\
PA-MPJPE & 37.1 & 36.0 & \textbf{33.9} & 49.8 & 47.2 \\
Rigid MPJPE & 43.7 & \textbf{43.5} & 54.8 & 59.8 & 56.2 \\
Instants scored/trial (angles) & 149 & 149 & 150 & 149 & 22 \\
Frames scored/trial (MPJPE) & 9003 & 9003 & 18631 & 8987 & 1285 \\
\bottomrule
\end{tabular}
\label{tab-movi-metrics-detail}
\end{table}

\begin{table}
\centering
\caption[]{Per-joint cross-method comparison on the clinical gait cohort against the MMMC reference. Joint-angle cells are MJAE (deg) with nIQR in parentheses; PA-MPJPE / Rigid MPJPE in mm. Best per row in bold; lower is better.}

\begin{tabular}{p{\dimexpr 0.167\linewidth-2\tabcolsep}p{\dimexpr 0.167\linewidth-2\tabcolsep}p{\dimexpr 0.167\linewidth-2\tabcolsep}p{\dimexpr 0.167\linewidth-2\tabcolsep}p{\dimexpr 0.167\linewidth-2\tabcolsep}p{\dimexpr 0.167\linewidth-2\tabcolsep}}
\toprule
Metric & Ours & Ours+IK & PBL† & HSMR & OpenCap† \\
\hline
\textit{Lower-limb joints} &  &  &  &  &  \\
Knee flexion & 4.33 (1.72) & 4.33 (1.73) & \textbf{3.99 (1.60)} & 17.70 (5.12) & 10.93 (6.89) \\
Hip flexion & 3.65 (1.63) & 3.60 (1.65) & \textbf{3.50 (1.84)} & 5.94 (3.45) & 9.58 (6.26) \\
Hip adduction & 2.30 (0.80) & 2.09 (0.80) & \textbf{1.66 (0.70)} & 3.52 (1.34) & 3.80 (2.11) \\
Hip rotation & 3.58 (1.02) & 3.53 (0.98) & \textbf{2.59 (0.92)} & 6.93 (3.02) & 5.90 (3.20) \\
Ankle flexion & 4.59 (1.63) & \textbf{4.58 (1.67)} & 5.13 (1.95) & 6.32 (1.73) & 9.02 (5.66) \\
\textit{Upper-limb joints} &  &  &  &  &  \\
Elbow flexion & 3.87 (1.84) & \textbf{3.81 (1.90)} & 4.19 (2.22) & 6.03 (3.81) & 6.22 (3.73) \\
Shoulder plane & 13.21 (5.56) & 13.29 (5.52) & \textbf{10.08 (6.09)} & 19.36 (8.23) & 27.48 (15.26) \\
Shoulder elevation & 3.62 (1.92) & 3.62 (1.87) & \textbf{2.00 (1.10)} & 4.26 (2.45) & 4.55 (2.58) \\
Shoulder rotation & 12.60 (5.19) & 12.61 (5.08) & \textbf{9.65 (5.25)} & 17.63 (8.25) & 29.31 (15.80) \\
\textit{Aggregate (pooled angle / MPJPE)} &  &  &  &  &  \\
Avg MJAE (°) & 5.75 & 5.72 & \textbf{4.75} & 9.74 & 11.87 \\
Avg MJAE, uncorrected (°) & 10.04 & 10.00 & 4.89 & 17.27 & 15.61 \\
PA-MPJPE & 41.9 & 42.0 & \textbf{27.6} & 75.8 & 82.4 \\
Rigid MPJPE & 50.4 & 50.5 & \textbf{27.2} & 82.6 & 90.3 \\
Instants scored/trial (angles) & 240 & 240 & 240 & 239 & 161 \\
Frames scored/trial (MPJPE) & 239 & 239 & 240 & 239 & 161 \\
Inputs rejected & 0 & 0 & 0 & 0 & 400 \\
\bottomrule
\end{tabular}
\label{tab-mmc-metrics-detail}
\end{table}

\subsection{A.2 Qualitative skeleton overlays}

Figure~\ref{fig-qual-mmmc-supp} and Figure~\ref{fig-qual-movi-supp} show the predicted
biomechanical skeleton of each method on the
clinical MMMC cohort and on held-out BML-MoVi respectively.

\begin{figure}[!htbp]
\centering
\includegraphics[width=1\linewidth]{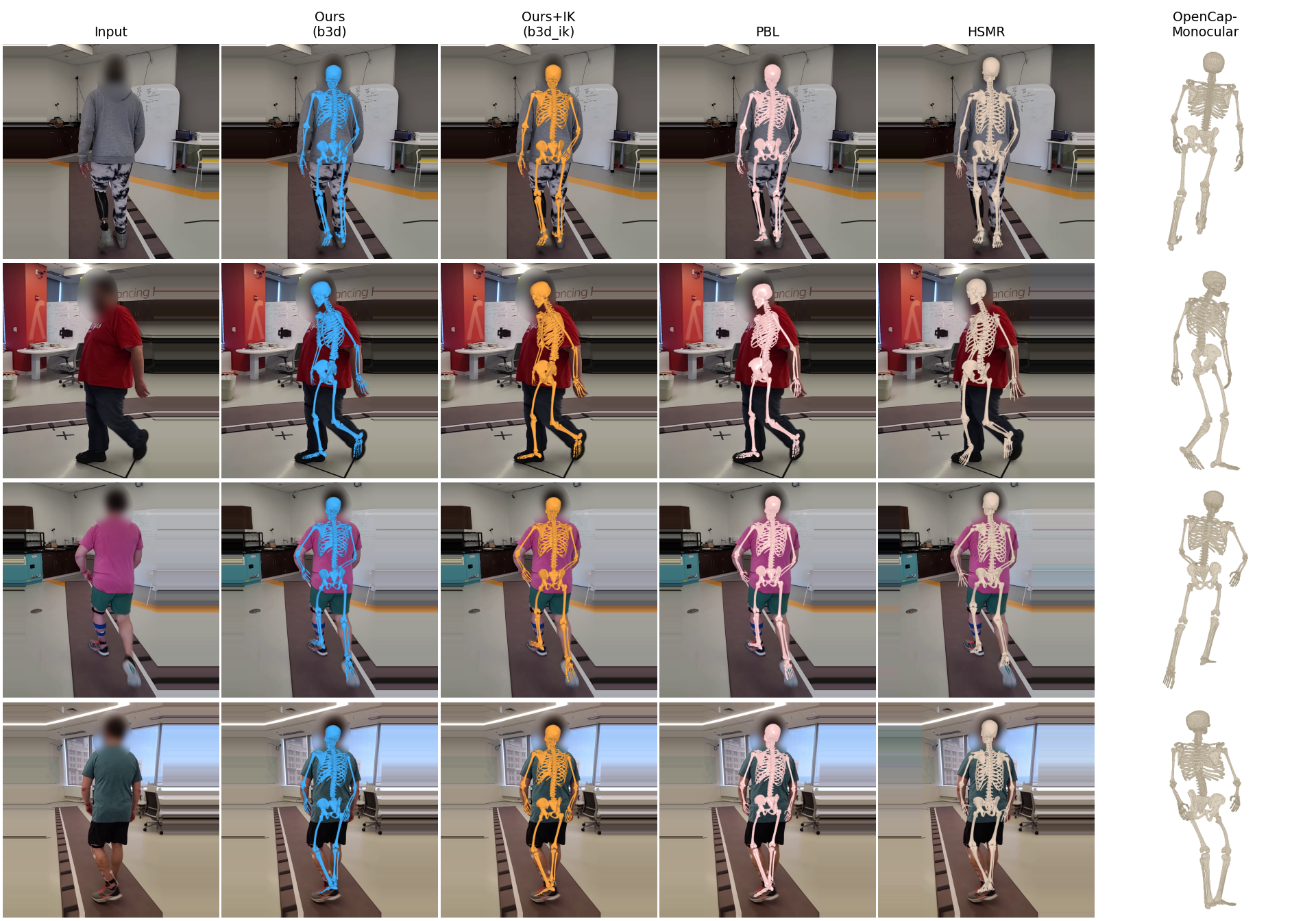}
\caption[]{Cross-method skeleton overlays on representative clinical-gait frames (MMMC cohort).
Columns: Input · Ours · Ours+IK · PBL · HSMR · OpenCap. Ours and PBL register the
lower limb tightly to the MMMC reference, while HSMR's SKEL skeleton carries a
visible knee/trunk offset.}
\label{fig-qual-mmmc-supp}
\end{figure}

\begin{figure}[!htbp]
\centering
\includegraphics[width=1\linewidth]{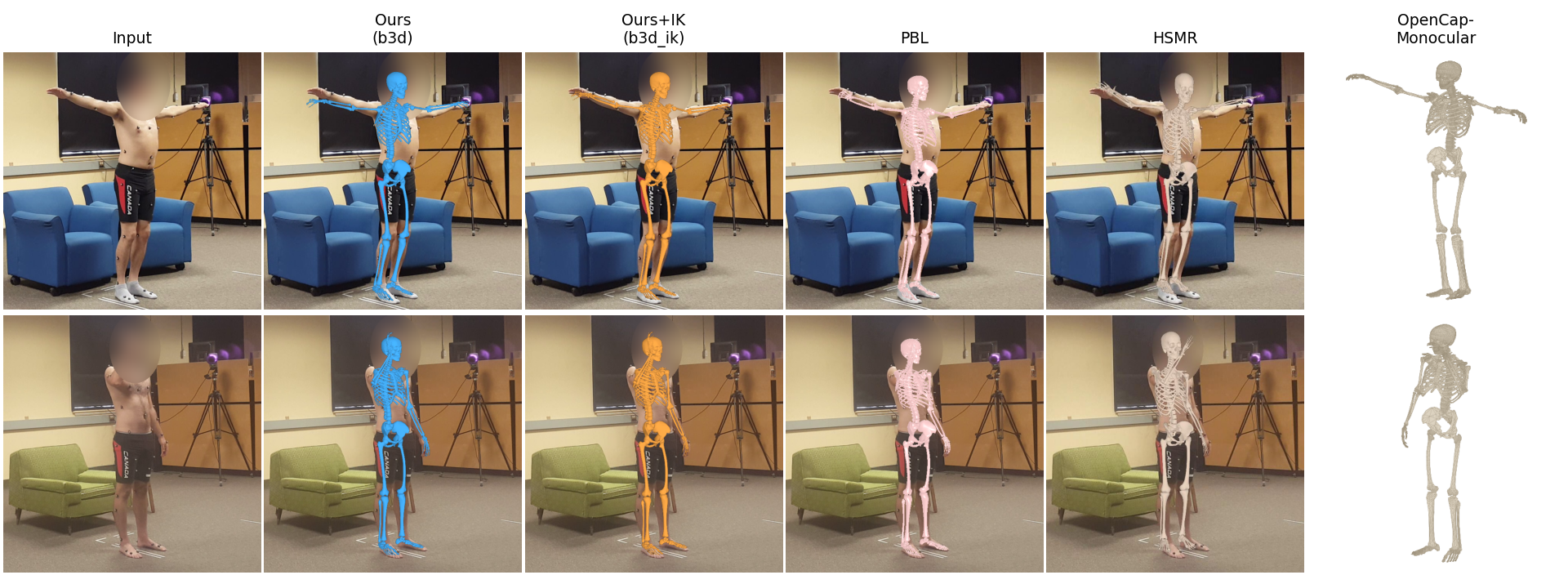}
\caption[]{Cross-method skeleton overlays on held-out BML-MoVi frames. Columns as above. On these well-observed lab frames Ours, Ours+IK, and PBL
register the limb segments closely to the optical-marker fit; HSMR's residual knee/trunk
offset persists even here, where the view is frontal and unoccluded.}
\label{fig-qual-movi-supp}
\end{figure}

\subsection{A.3 HSMR knee-flexion (crouch) bias}

Across the clinical MMMC cohort HSMR holds the knee in a permanently flexed,
forward-leaning \textit{crouch}, even on able-bodied control gait where the reference knee
reaches near-full extension at every stance phase (Figure~\ref{fig-hsmr-knee-bias}). Most
of the resulting error is a single constant offset --- a per-(method, joint) offset pooled
across all trials of \textbf{+30.4°} at the knee --- and every headline comparison is
bias-corrected, so this offset is removed before any method is compared and is not what
separates them. What a constant cannot fix is the per-frame shape, which is what the
figure shows: subtracting the offset aligns the waveform's mean but not its trajectory.

\begin{figure}[!htbp]
\centering
\includegraphics[width=0.75\linewidth]{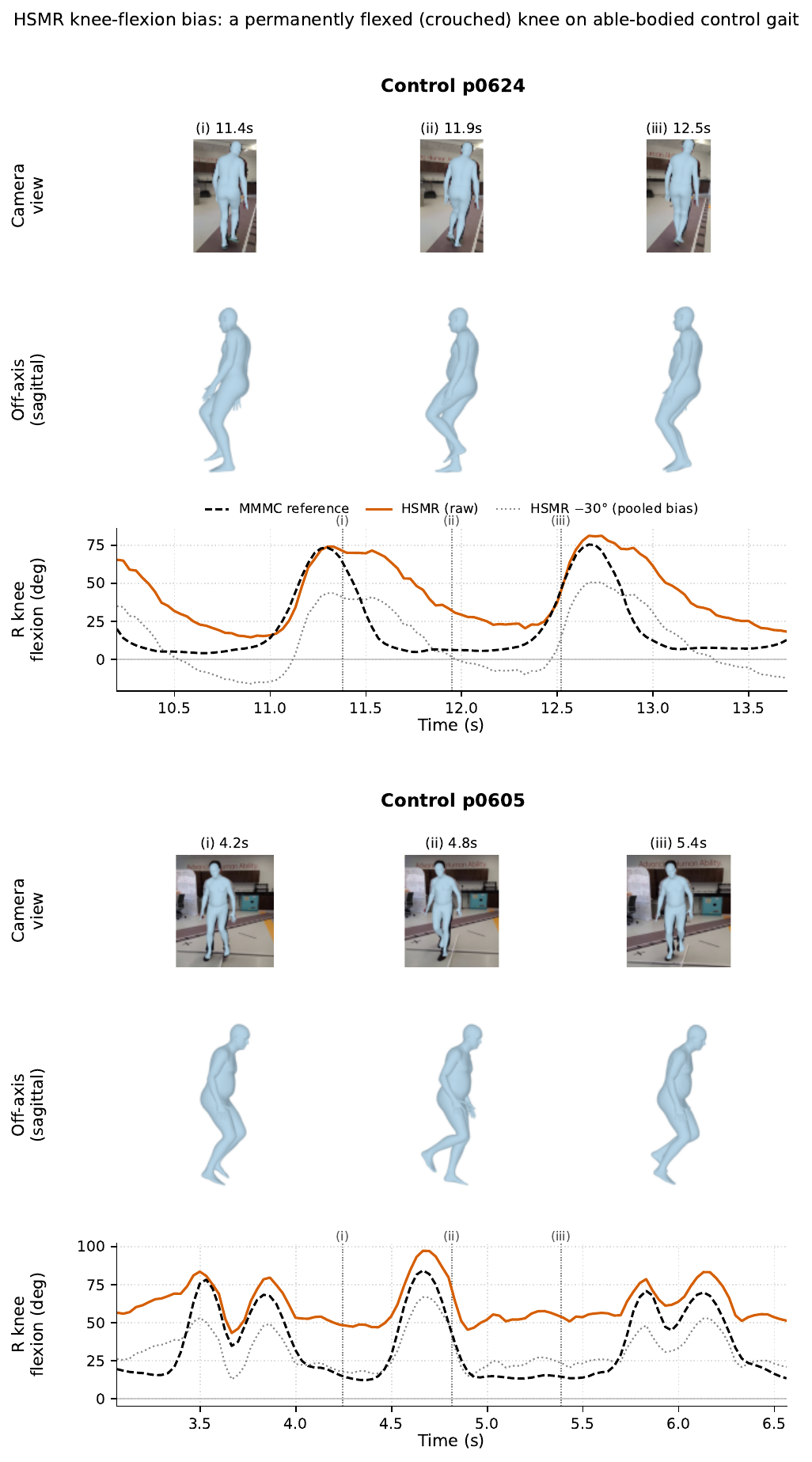}
\caption[]{HSMR knee-flexion bias on two able-bodied control gait trials (stacked). For each
trial: the HSMR skeleton mesh shown image-registered on the camera frame (top row)
and from an off-axis sagittal camera (middle row) at three sampled gait phases
(i, ii, iii), over the right-knee-flexion waveform (bottom). Black dashed = the
MMMC reference; orange = raw HSMR; faint dotted = HSMR after
subtracting the single pooled +30.4° knee bias. Vertical lines mark the time of each
sampled mesh. The off-axis profiles reveal the permanently flexed (crouched) knee
that produces the bias; the constant offset aligns the waveform mean but not its
per-frame shape.}
\label{fig-hsmr-knee-bias}
\end{figure}

\subsection{A.4 Why the MHR head is kept frozen: deceptive in-distribution metrics}

Figure~\ref{fig-finetune-dynamics} shows training dynamics from an experiment
that fine-tuned the MHR head during distillation. Every in-distribution curve improves
monotonically, yet off-axis viewpoints show the recovered mesh collapses.
Figure~\ref{fig-finetune-collapse} shows that the fine-tuned mesh
looks plausible frontally, but rotated off-axis it folds and twists.
The in-distribution metrics give no warning, so we keep the MHR head frozen.

\begin{figure}[!htbp]
\centering
\includegraphics[width=0.9\linewidth]{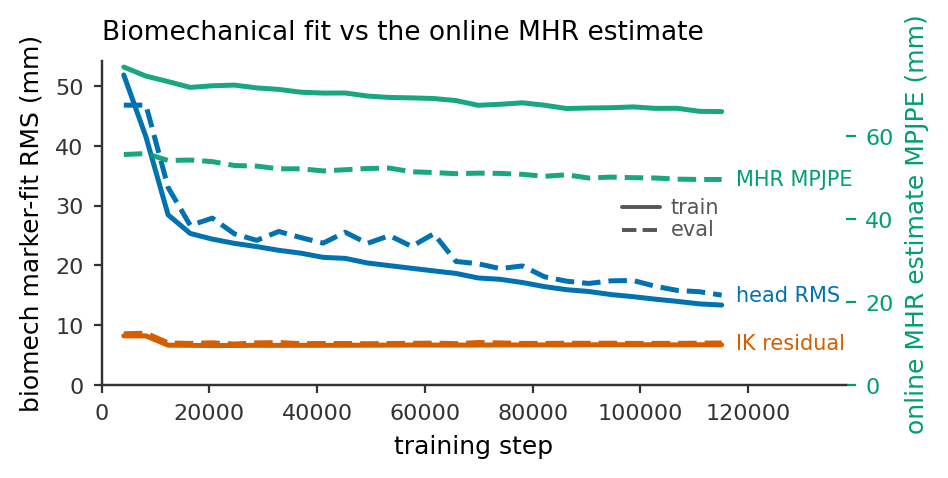}
\caption[]{Training dynamics of the discarded MHR-head fine-tuning run (train solid, eval
dashed). Online MHR keypoint MPJPE (green), biomechanical marker RMS (blue), and
inline-IK residual (orange) all improve in-distribution throughout training, while the
mesh degrades catastrophically on off-axis views.}
\label{fig-finetune-dynamics}
\end{figure}

\begin{figure}[!htbp]
\centering
\includegraphics[width=1\linewidth]{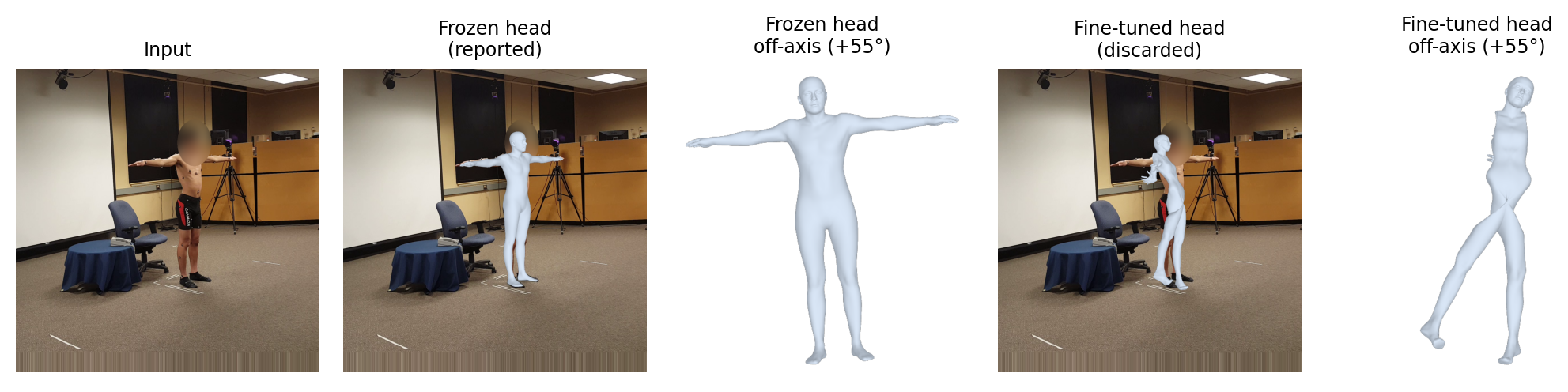}
\caption[]{Frozen vs fine-tuned MHR mesh on the worst-case (maximum frozen-vs-fine-tuned
vertex-divergence) held-out BML-MoVi frame. Columns: input · frozen
(frontal) · frozen (+55°) · fine-tuned (frontal) · fine-tuned (+55°). Both look
plausible frontally; off-axis the fine-tuned mesh collapses while the frozen mesh
stays anatomical. This frame illustrates the failure mode at its most severe; the
typical off-axis drift of the fine-tuned head is smaller but still substantial.}
\label{fig-finetune-collapse}
\end{figure}

\subsection{A.5 What the biomechanical head adds over IK alone}

The head is trained to reproduce what the inline solver computes, so a natural
question is whether it earns its place: the same Levenberg--Marquardt IK could simply
be run on the frozen MHR output at inference, with no learned head at all. We test
this directly by scoring a third variant on BML-MoVi that runs the identical solver
seeded from the neutral pose rather than from the head's prediction, leaving the
head bypassed entirely.

On typical frames the two are close; the difference is in the tail, where the
cold-started solver produces substantially more outlier frames
(Table~\ref{tab-coldik-ablation}). This aligns with the broader difficulty of inverse kinematics on a kinematic chain: with the marker offsets
themselves free, the fit is both computationally demanding and prone to settling into a
local minimum that explains the markers while getting the pose wrong. The head supplies
a consistent, image-conditioned starting point that keeps the solve in the right basin,
at a fraction of the cost.

\begin{table}
\centering
\caption[]{Feed-forward head vs cold-started IK on the frozen MHR output, BML-MoVi. All three variants are scored on the same subjects, at the same instants, and with the same pooled per-(method, joint) bias correction as the main tables. The joint-angle rows are computed at the frames where the network actually ran; the PA-MPJPE rows are computed over each variant's full resampled series, identically for all three. Lower is better throughout.}

\begin{tabular}{p{\dimexpr 0.250\linewidth-2\tabcolsep}p{\dimexpr 0.250\linewidth-2\tabcolsep}p{\dimexpr 0.250\linewidth-2\tabcolsep}p{\dimexpr 0.250\linewidth-2\tabcolsep}}
\toprule
Metric (BML-MoVi, n=85) & Ours & Ours+IK & MHR + cold IK \\
\hline
Lower-limb MJAE (deg) & 3.70 (1.35) & 3.48 (1.45) & 3.60 (1.41) \\
Upper-limb MJAE (deg) & 7.39 (2.16) & 7.38 (2.38) & 7.90 (2.46) \\
PA-MPJPE, median (mm) & 37.2 & 36.4 & 37.9 \\
PA-MPJPE, mean (mm) & 38.1 & 37.5 & 39.2 \\
PA-MPJPE, p90 (mm) & 45.1 & 44.9 & 47.5 \\
Lower-limb error spread p95 (deg) & 9.25 & 9.13 & 11.78 \\
Lower-limb error spread p99 (deg) & 15.11 & 15.18 & 26.09 \\
Frames \textgreater 15 deg from trial median (\%) & 1.07 & 1.08 & 3.04 \\
Frames \textgreater 30 deg from trial median (\%) & 0.00 & 0.00 & 0.80 \\
Inference (s/frame) & 2.73 & 2.73 & 5.06 \\
Relative cost, in harness & 1.0x & 1.0x & 1.9x \\
\bottomrule
\end{tabular}

The per-frame times are wall-clock inside the evaluation harness, which decodes video and pads every chunk to a fixed batch, so they are not a clean measure of the solver alone: at this dataset's frame sampling each padded chunk carries about one scored frame, so the cold solve runs essentially unbatched. They bound the cost as evaluated here (2.3 s/frame added over the head) rather than the cost a batched deployment would pay. See the standalone benchmark for a component-level split of forward pass versus IK.
\label{tab-coldik-ablation}
\end{table}

\subsection{A.6 Per-method gait waveforms across trials}

Figure~\ref{fig-gait-waveforms-supp} shows hip and knee kinematic waveforms for two clinical walking trials. Unlike the raw main-paper overlay, these supplementary waveforms are bias-corrected
(each method's single pooled per-joint offset removed), so they isolate waveform
\textit{shape} rather than constant convention offset. These qualitative results align with our quantitative results that PBL and B3D show the best performance.

\begin{figure}[!htbp]
\centering
\includegraphics[width=1\linewidth]{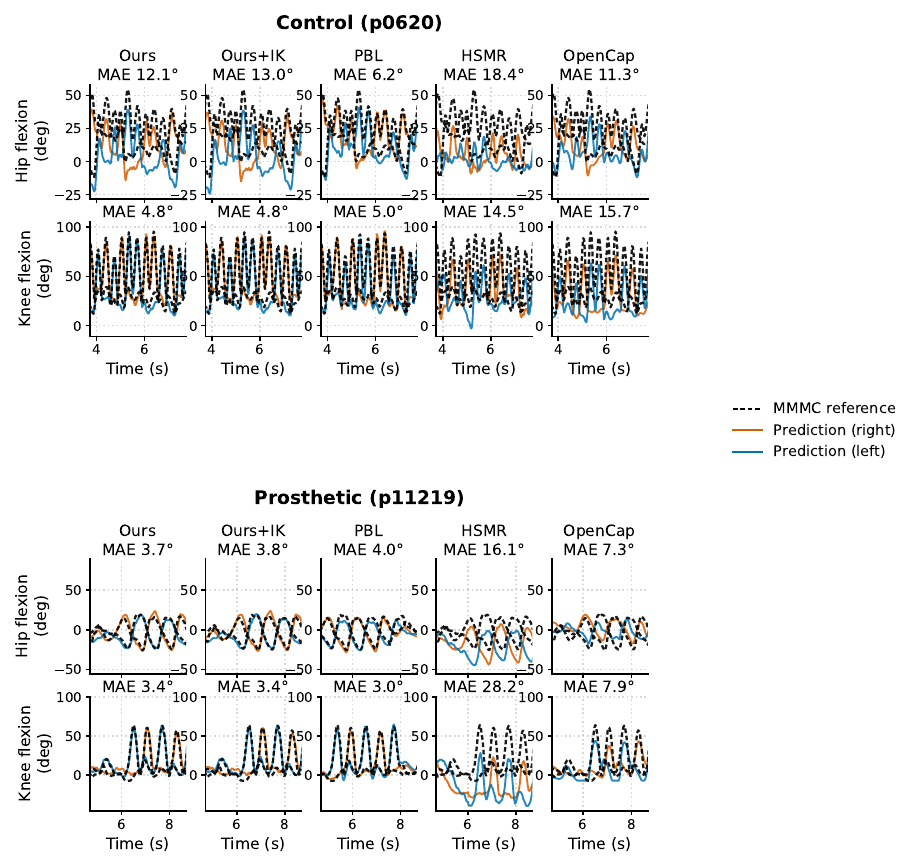}
\caption[]{Per-method gait waveforms on two clinical walking trials (one control and one
lower-limb prosthesis user; hip above knee per trial; columns Ours / Ours+IK / PBL / HSMR /
OpenCap), 4 s windows. Black dashed: sign-corrected reference; red and blue:
predicted right and left; predictions bias-corrected as in the result tables. Panel
titles give the per-leg-averaged error. HSMR's knee decorrelates on the prosthetic
trial while Ours and PBL stay close to the reference.}
\label{fig-gait-waveforms-supp}
\end{figure}

\section{Methods, metrics, and training detail}

This appendix records how the model was trained and evaluated. It describes how the distillation dataset was assembled, the training schedule and loss weights, the metric conventions applied throughout, and a check confirming that the JAX/Equinox port stays faithful to the reference implementation.

\subsection{A.7 Training dataset detail}

We distill from the public SAM-3D-Body dataset \parencite{yang_sam_2026}, released on
HuggingFace as \texttt{facebook/sam-3d-body-dataset} \parencite{sam3dbody_dataset_hf}. The
dataset does not provide its own images; it supplies SAM-3D-Body / MHR annotations
layered on top of eight publicly available image sources. Our training index uses
\textbf{five} of these --- SA-1B, Harmony4D, COCO, 3DPW, and MPII; the
remaining three (AI Challenger, EgoExo4D, EgoHumans) are not included
(Table~\ref{tab-train-sources} gives the per-source image and person-crop counts). We use it purely as a source of
paired images and MHR outputs for distillation, applying only mild colour and
bounding-box jitter (no flips or rotations) so the geometric correspondence between
image and target is preserved.

\begin{table}
\centering
\caption[]{Per-source composition of the SAM-3D-Body training data. The dataset layers MHR annotations on eight public image sources; our training index draws from five. Counts are from the training index: ``person-crops'' are individual bounding-box training samples, ``images'' are distinct source images. The held-out evaluation split (195,696 crops / 78,249 images) is excluded.}
\label{tab-train-sources}
\begin{tabular}{p{\dimexpr 0.250\linewidth-2\tabcolsep}p{\dimexpr 0.250\linewidth-2\tabcolsep}p{\dimexpr 0.250\linewidth-2\tabcolsep}p{\dimexpr 0.250\linewidth-2\tabcolsep}}
\toprule
Image source & Distinct images & Person-crops & Used \\
\hline
SA-1B & 1,206,635 & 3,185,516 & yes \\
Harmony4D & 235,831 & 471,662 & yes \\
COCO & 22,589 & 37,632 & yes \\
3DPW & 16,508 & 21,232 & yes \\
MPII & 5,157 & 6,734 & yes \\
AI Challenger & --- & --- & no \\
EgoExo4D & --- & --- & no \\
EgoHumans & --- & --- & no \\
\textbf{Total (used)} & \textbf{1,486,720} & \textbf{3,722,776} &  \\
\bottomrule
\end{tabular}
\end{table}

\subsection{A.8 Training schedule, loss weights, and $\mathcal{R}(q,\beta)$}

The reported model was trained for 35 epochs; the values below are its exact
configured hyperparameters.

\textbf{Optimization.} 35 epochs $\times\,4{,}111$ steps/epoch $= 143{,}885$ total
optimizer steps; batch size 16; learning rate $3\times10^{ -4}$ with linear warmup
over 50 steps and cosine decay to a floor ratio of 10\textsuperscript{-3}; seed 42. The MHR
head is frozen throughout. Training ran for $\approx$134 hours ($\approx$5.6 days)
of wall-clock time on a single NVIDIA L40S GPU (the inline Levenberg--Marquardt-IK inner solve, run
every step, dominates the per-step cost).

\textbf{Loss-term weights.} The training objective (the loss defined in the main text)
decomposes into the distillation residual $r$, the inline-IK pseudo-label terms, and
the regularizer $\mathcal{R}(q,\beta)$:

\begin{table}
\centering
\caption[]{Loss-term weights for the reported run, grouped by their role in the main-text training objective. Config names are the training flags.}
\label{tab-loss-weights}
\begin{tabular}{p{\dimexpr 0.250\linewidth-2\tabcolsep}p{\dimexpr 0.250\linewidth-2\tabcolsep}p{\dimexpr 0.250\linewidth-2\tabcolsep}p{\dimexpr 0.250\linewidth-2\tabcolsep}}
\toprule
Group & Term & Flag & Weight \\
\hline
Distillation $r$ & marker residual & \texttt{w\_marker} & 1.0 \\
IK pseudo-labels & joint angles $q$ & \texttt{w\_joint} & 2.5 \\
 & root pose & \texttt{w\_root} & 2.5 \\
 & segment scale $\beta$ & \texttt{w\_scale} & 1.0 \\
 & marker offsets & \texttt{w\_offset\_sup} & 10.0 \\
Regularizer $\mathcal{R}$ & joint-limit + ball-cone & \texttt{w\_joint\_limits} & 100.0 \\
 & 6D column-norm prior & \texttt{w\_quat\_reg} & 0.1 \\
 & scale prior & \texttt{w\_scale\_reg} & 0.05 \\
 & offset prior (inactive) & \texttt{w\_offset\_reg} & 0.0 \\
 & FK scale-clip range & \texttt{fk\_scale\_clip} & $[0.05, 5.0]$ \\
\bottomrule
\end{tabular}
\end{table}

Mapping these to the main-text objective (Eq. 4 of the main paper): the distillation weight
$w_\text{r}$ is \texttt{w\_marker} $=1.0$; the coordinate weight $w_q$ is \texttt{w\_joint} $=2.5$
(with a separate root sub-term \texttt{w\_root} $=2.5$); and the shape weight $w_\beta$
decomposes into \texttt{w\_scale} $=1.0$ for the per-segment scales and \texttt{w\_offset\_sup}
$=10.0$ for the marker offsets (recall $\beta$ bundles both).

\textbf{Terms of $\mathcal{R}(q,\beta)$.} The regularizer collects: (1) a \textbf{joint-limit
penalty} on the hinge/slide DOFs (mean squared violation of the anatomical lower/upper
bounds); (2) a \textbf{ball-cone penalty} on the ball joints (hip, shoulder) penalizing
rotation beyond the allowed cone half-angle; (3) a \textbf{6D column-norm prior} pulling the
two predicted 6D-rotation columns of each joint toward unit norm (the rotation
representation is 6D, \texttt{use\_6d}); (4) a \textbf{scale prior} pulling the per-segment scales
toward 1; and (5) \textbf{FK scale clipping}, a hard clamp of every per-body scale
multiplier inside FK (range in Table~\ref{tab-loss-weights}; gradient zeroed on saturation),
which suppresses a self-distillation scale runaway. An \textbf{offset-magnitude prior}
(\texttt{w\_offset\_reg}) is available but disabled (weight 0) in the reported run; offset
shrinkage is instead supplied through the IK pseudo-label path (\texttt{w\_offset\_sup}).

\textbf{Inline Levenberg--Marquardt-IK staging and the cold-to-warm transition.} The inline solver runs three
Levenberg--Marquardt stages each step --- (i) root, (ii) pose $+$ scale, (iii) full pose
$+$ marker offsets --- with damping 2.0 (a fourth hand-refinement stage is disabled, as
the hand pose is not refined or evaluated here). For the first \texttt{cold\_warm\_step} $=8{,}000$ optimizer steps the solver
is \textbf{cold-started} from the neutral pose with $[133, 133, 134] = 400$ Levenberg--Marquardt iterations
across the three stages; once training passes $8{,}000$ steps --- at the epoch $2\to3$
boundary, given $4{,}111$ steps/epoch --- it switches to a \textbf{warm start} seeded from the
head's own biomechanical prediction with $[100, 100, 100] = 300$ iterations. The warm
start escapes the poor local minima a neutral seed settles into on the highly coupled
kinematic tree (training-dynamics figure, main text).

\subsection{A.9 Scale groups and the scaling matrix}

Rather than scaling each segment independently, we group anatomically related segments
and predict one scalar per group. The \textbf{scaling matrix} maps these group scalars to the
per-body scale vector used by FK, enforcing left/right symmetry and
keeping the scale parameterisation low-dimensional.

There are 14 groups: trunk (pelvis + head), thigh/shin/foot per leg, upper-arm/forearm/hand per arm, and one overall body scale. The hands have their own dedicated groups so hand-marker scale mismatches are absorbed independently from the arm. Group scalars are regularized toward 1 and hard-clamped inside FK (see Table~\ref{tab-loss-weights}).

\subsection{A.10 Metric conventions: bias correction and the 19-joint set}

Bias-corrected angle metrics subtract one fixed per-(method, joint-group) offset before differencing --- a single value per method, pooled over the whole cohort rather than fit per participant, so it removes only each skeleton's constant convention or frame offset and cannot absorb genuine per-subject pose error. The same correction is applied identically to every method, so no method is advantaged; uncorrected rows report each method's raw per-group mean for reference.

Each joint center used in PA-MPJPE is the world-frame position of a named body's local reference frame origin after FK. For example, the femur body's frame origin serves as the hip joint center, the tibia body's as the knee joint center, and so on, following the anatomical placement built into each model's kinematic tree during its original construction. PA-MPJPE pools the 19 such centers common to every method: the pelvis, and bilaterally the femur, tibia, talus, calcaneus and toes in the lower limb and the humerus, ulna, radius and hand in the upper limb. The torso and head are excluded because not all baselines expose them, and the fingers are excluded because finger pose is not fitted here. The per-frame similarity transform (scale, rotation, translation) is fit using each method's full exposed joint set, not just the 19 joint centers, and the resulting residual is then averaged over the 19 common centers. Restricting the fit itself to the 19 centers changes the score by 0.16 mm on average and at most 0.35 mm.

The feed-forward methods (ours, HSMR) run per-frame and are scored directly on the
selected reference frames. OpenCap and PBL are not per-frame methods. OpenCap runs WHAM
(a temporal 3D-pose model) followed by OpenSim scaling and IK over a
contiguous person-cropped clip spanning the subject's full in-view presence (clamped to
600 source frames by GPU memory), producing a single full-video trajectory that is then
linearly resampled onto the reference frame grid for scoring; reference frames outside
that trajectory's support are excluded with no extrapolation. WHAM's native keypoint gate
($\geq$21 of 25 OpenPose keypoints at confidence $\geq$0.8) is the final arbiter of which frames
survive, so OpenCap's actually-compared frame set is a subset of the shared grid, and a
fraction of trials are dropped entirely where it does not produce a usable trajectory.
On the clinical cohort this gate rejected 400 of 1693 inputs --- the posterior and
self-occluded views common in our data often fail the keypoint-visibility check ---
consistent with OpenCap-Monocular being validated for a 45° anterolateral view
\parencite{gilon_opencap_2026}; excluding these may bias its reported accuracy favorably.
PBL is a precomputed whole-clip monocular optimization fit, re-scored here by sampling
its stored trajectory at the reference frames.

\subsubsection{A.10.1 BioCV evaluation cohort: synchronisation and quality gating}

BioCV \parencite{evans_biocv_2024} is a University of Bath dataset in which a single
acquisition trigger hardware-synchronises nine 200 Hz cameras with a Qualisys optical
marker system, so video frame $n$ and marker frame $n$ are the same instant with no
software time-alignment --- the strongest synchronisation guarantee of our three
references. We run monocular inference from a single anterolateral camera (cam01) with a
tight per-frame person-detection crop and score against a marker-based MuJoCo implicit
kinematic reconstruction across walking (129), running (81), self-selected and maximal
effort counter-movement jumps (45 + 44), and hopping (9).

From the 362 recorded trials we exclude static poses (not a movement task), 43
trials whose reference kinematic fits failed to converge --- identified by the
reference fit's own median marker residual exceeding 50 mm, which cleanly separates
the converged references ($<20$ mm) from the diverged ones ($>250$ mm) across an empty
gap --- and 3 trials for which the subject was never tracked within the analysis window
(in-window bounding-box coverage below 25\%), leaving the monocular front-end with no
valid frames. Every exclusion is applied identically to all methods so it cannot bias
the comparison, leaving 308 scored trials. The reference-health gate thresholds a
median, so a trial whose reference degrades only briefly survives it: four of the
retained trials carry a 90th-percentile marker residual between 50 and 66 mm against
a typical retained trial's 13 mm. Excluding all four shifts every method's joint-angle
rows by at most 0.09° and its PA-MPJPE by 0.2--0.7 mm, in the same direction for every
column, so we report the full 308.

The PBL column on BioCV is the lab's single-camera whole-clip optimisation fit of the
same cam01 video (2D keypoints plus a monocular 3D lift; no marker input), so it is a
monocular baseline on the same footing as the other columns. It covers 267 of the 308
trials: two participants were never fit at cam01, and their trials therefore carry no PBL
value. We deliberately do not report the dataset's marker-based optimisation fit in this
column --- it is fit to the very Qualisys markers the reference is built from, so scoring it
against that reference measures agreement between two fits of the same input rather than
monocular accuracy.

\subsubsection{A.10.2 Sampling density, per-trial caps, and the matched scoring instants}

The methods do not observe a trial equally densely, and the joint-angle metrics are
computed so that this does not become an accuracy difference.

The feed-forward methods (ours, HSMR) pass every scored frame through a network, so for
inference compute we subsample the high-rate datasets rather than running every frame.
On BML-MoVi we take every 15th video frame --- 2 Hz against a 120 Hz marker reference ---
and cap each trial at 150 such frames, which is the binding limit; on BioCV, every
20th frame of the 200 Hz capture, i.e. 10 Hz, with the cap never reached. On the
clinical cohort, whose videos are 30 Hz, no subsampling is applied and the median trial
contributes 240 frames. Both
feed-forward methods use the same sampling on the same trials.

The two optimization methods run at the full frame rate by design and are not
subsampled: PBL fits an entire clip in one optimization, and OpenCap's WHAM front-end
runs per video frame. Their per-trial limits are therefore set by different constraints
than ours --- OpenCap processes a contiguous in-view window clamped to 600 source frames
by GPU memory --- so the caps themselves are \textit{not} matched across methods, and we do not
claim they are.

What is matched is where the comparison is evaluated. Every method is scored at the
same instants: those at which the subsampled feed-forward methods actually ran,
recovered from their stored per-frame outputs. Each method contributes its own value at
those instants, and a dense method loses nothing by it, since it has a real estimate
there too. Without this step roughly 95\% of the frames the feed-forward error was
computed over would be linear interpolations between predictions up to half a second
apart, while the optimization methods were scored almost entirely on real estimates ---
so the columns would answer two different questions, one about the reconstructed
trajectory and one about the estimate at an instant.

This equalizes sampling density, not temporal coverage. A method is matched only where
its own reconstruction exists, so one that covers part of a trial is scored on that
part: OpenCap's BML-MoVi trajectories span about 10.7 s of the {\textasciitilde}75 s scored window and
reach a median of 22 of the 149 instants. The realized count per method is reported in the
per-joint tables of Appendix A.1.

The joint-centre metrics (PA-MPJPE, rigid MPJPE) are the exception and keep the older
basis, scored over each method's full resampled series. Not every baseline retains
per-frame 3D joint positions locally, so those rows cannot all be re-cut to the matched
instants; they are computed identically for every method, and the tables report their
frame count on a separate row so the two bases are not confused.

\subsection{A.11 Port faithfulness --- JAX/Equinox vs the PyTorch teacher}

Because our biomechanical head distills from a frozen SAM-3D-Body trunk that we
re-implemented in JAX/Equinox (for a single MJX-differentiable graph), we first
establish that the port itself is not a source of error. We run the Meta
\textbf{PyTorch SAM-3D-Body} model and our \textbf{JAX/Equinox port} on the SAME samples
drawn from the SAM-3D-Body \textit{dataset} (a seeded permutation over the full eval
index, spanning SA-1B / Harmony4D / COCO / 3DPW; n=250), and score BOTH with the
identical training metric (MPJPE / PA-MPJPE / PCK over the 70 MHR joints,
pelvis-rooted, full-skeleton Procrustes for PA). The
ground truth is the dataset pseudo-GT (the Meta model's stored fits).

The raw torch-converted trunk --- before any distillation --- is scored against the PyTorch teacher on the same 250 crops (Table~\ref{tab-port-parity}). The port costs only \textbf{+0.08 mm MPJPE / +0.13 mm PA-MPJPE} (direct PT$\leftrightarrow$JAX disagreement 3.42 mm over all 70 joints, $\approx$2.1 mm after Procrustes) --- two orders of magnitude below the biomechanical errors in the main-text comparison. The reimplementation is faithful and contributes negligibly to reconstruction error.

\begin{table}
\centering
\caption[]{PyTorch vs JAX/Equinox port parity for the SAM-3D-Body trunk (n=250 crops, pre-distillation). MPJPE/PA-MPJPE are pelvis-rooted over the 70 MHR joints; PCK is 2D normalised by bounding-box diagonal. $\Delta$ = JAX - PyTorch; lower magnitude is better.}
\label{tab-port-parity}
\begin{tabular}{p{\dimexpr 0.250\linewidth-2\tabcolsep}p{\dimexpr 0.250\linewidth-2\tabcolsep}p{\dimexpr 0.250\linewidth-2\tabcolsep}p{\dimexpr 0.250\linewidth-2\tabcolsep}}
\toprule
Metric & PyTorch & JAX/Equinox port & $\Delta$ (port cost) \\
\hline
MPJPE (mm) & 64.32 & 64.40 & +0.08 \\
PA-MPJPE (mm) & 43.55 & 43.68 & +0.13 \\
PCK @ 0.01 & 0.3645 & 0.3625 & -0.0020 \\
PCK @ 0.025 & 0.6982 & 0.6971 & -0.0011 \\
PCK @ 0.05 & 0.8201 & 0.8199 & -0.0002 \\
PCK @ 0.075 & 0.8574 & 0.8578 & +0.0004 \\
PCK @ 0.1 & 0.8799 & 0.8804 & +0.0005 \\
PCK (avg) & 0.7240 & 0.7235 & -0.0005 \\
\bottomrule
\end{tabular}
\end{table}

\subsection{A.12 Biomechanical head architecture}

The trainable biomechanical head is a lightweight ({\textasciitilde}13.6 M-parameter) module that
reads a single 1024-d pose token from the frozen SAM-3D-Body decoder. With MHR
conditioning enabled, the frozen MHR head's output (body pose and camera, 136-d)
is mapped to token space by a learned linear layer and concatenated with the pose
token (2048-d input), so the head's task reduces to remapping the MHR prediction
into MuJoCo coordinates. Three parallel two-layer MLPs (hidden width 2048, ReLU,
no residual or normalization) produce the three outputs. The generalized
coordinates are emitted in a 109-d network space that uses 6D continuous rotations
\parencite{zhou_6d_2019} for the root free joint and the four hip/shoulder ball
joints; a Gram--Schmidt step maps each 6D block to a unit quaternion, giving the
99-d physical MuJoCo \texttt{qpos}, while hinge joints pass through as scalar angles.
Scale clipping and the offset prior act downstream in FK (§A.8),
not inside the head.

\begin{table}
\centering
\caption[]{Biomechanical-head outputs. The three regressors are independent two-layer MLPs off the shared 2048-d input; the only extra parameters are the MHR-conditioning projection.}
\label{tab-head-arch}
\begin{tabular}{p{\dimexpr 0.333\linewidth-2\tabcolsep}p{\dimexpr 0.333\linewidth-2\tabcolsep}p{\dimexpr 0.333\linewidth-2\tabcolsep}}
\toprule
Regressor & Output & Dim \\
\hline
Generalized coordinates & 6D rotations (root + 4 ball joints) + hinge scalars & 109 (99 physical) \\
Per-segment scale & zero-centerd scale deltas & 14 \\
Marker offsets & per-marker XYZ, metres & 306 (102$\times$3) \\
\bottomrule
\end{tabular}
\end{table}

\end{document}